\documentclass[letterpaper, 10 pt, conference]{ieeeconf}  
\usepackage{times}
\usepackage[utf8]{inputenc} 
\usepackage[T1]{fontenc}    
\usepackage{url}            
\usepackage{booktabs}       
\usepackage{nicefrac}       
\usepackage{microtype}      
\usepackage{adjustbox}      
\usepackage{subcaption}
\usepackage{wrapfig}
\usepackage[font=footnotesize,labelfont=bf]{caption}

\usepackage{algorithm}
\usepackage[noend]{algpseudocode}
\algrenewcommand\algorithmicindent{1em}
\algrenewcommand{\algorithmiccomment}[1]{%
\bgroup\hskip2em\textcolor{ourdarkgreen}{//~\textsl{#1}}\egroup}

\let\labelindent\relax
\usepackage{enumitem}

\usepackage{amsfonts}       
\usepackage{amsmath}       
\usepackage{amssymb}
\usepackage{xspace}
\usepackage{multirow}

\usepackage{xr-hyper}
\usepackage{hyperref}
\usepackage[capitalise]{cleveref} 

\makeatletter
\DeclareRobustCommand\onedot{\futurelet\@let@token\@onedot}
\def\@onedot{\ifx\@let@token.\else.\null\fi\xspace}
\makeatother

\newcommand{\ie}{i.e\onedot}

\newcommand{\RNum}[1]{\uppercase\expandafter{\romannumeral #1\relax}}

\def\eqref#1{(\ref{#1})}

\makeatletter
\newcommand*{\addFileDependency}[1]{
  \typeout{(#1)}
  \@addtofilelist{#1}
  \IfFileExists{#1}{}{\typeout{No file #1.}}
}
\makeatother

\usepackage{xcolor}
\definecolor{ourblue}{rgb}{0.368,0.507,0.71}
\definecolor{ourorange}{rgb}{0.881,0.611,0.142}
\definecolor{ourgreen}{rgb}{0.56,0.692,0.195}
\definecolor{ourred}{rgb}{0.923,0.386,0.209}
\definecolor{ourviolet}{rgb}{0.528,0.471,0.701}
\definecolor{ourbrown}{rgb}{0.772,0.432,0.102}
\definecolor{ourlightblue}{rgb}{0.364,0.619,0.782}
\definecolor{ourdarkgreen}{rgb}{0.572,0.586,0.}

\definecolor{ourcyan2}{rgb}{0.125,0.722,0.804}
\definecolor{ourred2}{rgb}{0.863,0.184,0.047}
\definecolor{ouryellow2}{cmyk}{0,0.16,1.0,0.07}
\definecolor{ourviolet2}{cmyk}{0.55,0.56,0,0.47}
\definecolor{ourorange2}{cmyk}{0,0.46,0.89,0.11}

\IEEEoverridecommandlockouts                              

\title{\LARGE \bf
The Role of Tactile Sensing for Learning Reach and Grasp
}

\author{Boya Zhang$^{1}$, Iris Andrussow$^{1,2}$, Andreas Zell$^{1}$ and Georg Martius$^{1,2}$
\thanks{$^{1}$Department of Computer Science, University of Tübingen, Germany
        {\tt\small firstname.surname@uni-tuebingen.de}}%
\thanks{$^{2}$Max Planck Institute for Intelligent Systems, Germany
        {\tt\small surname@is.mpg.de}}%
}

\begin{document}

\maketitle
\thispagestyle{empty}
\pagestyle{empty}

\begin{abstract}
Stable and robust robotic grasping is essential for current and future robot applications. In recent works, the use of large datasets and supervised learning has enhanced speed and precision in antipodal grasping. However, these methods struggle with perception and calibration errors due to large planning horizons. To obtain more robust and reactive grasping motions, leveraging reinforcement learning combined with tactile sensing is a promising direction. Yet, there is no systematic evaluation of how the complexity of force-based tactile sensing affects the learning behavior for grasping tasks. This paper compares various tactile and environmental setups using two model-free reinforcement learning approaches for antipodal grasping. Our findings suggest that under imperfect visual perception, various tactile features improve learning outcomes, while complex tactile inputs complicate training.

\end{abstract}

\keywords
Robotics, Tactile Sensing, Grasping, Reinforcement Learning
\endkeywords
\section{Introduction}\label{sec:intro}
Reactive grasping is the foundation for robust interaction of a robot with its environment. Multiple deep reinforcement learning-based grasping methods have been proposed to learn reactive and efficient grasping policies \cite{kalashnikov2018scalable,quillen2018deep,pavlichenko2023deep,koenig2022role}.
Besides visual information from the camera system and proprioception, tactile sensors on the gripper can provide additional information about contact to improve the learning of grasping behaviors and their robustness.
However, the multi-modal nature and high bandwidth of tactile information pose challenges to learning-based methods. Multiple tactile sensors with force output have been developed for robotic manipulation. They differ mainly in their sensing area, sensing resolution and the quantities they report. 
\emph{But which sensor type is most suitable for grasping with reinforcement learning?}
We study this question by modeling a prototypical shape of a robotic fingertip and simulating various types of tactile information while learning grasping behaviors using a parallelized simulation environment. We evaluated our results in real robot experiments, using the Minsight sensor \cite{Andrussow23-AIS-Minsight} as it allows emulating a wide range of tactile sensors.

Our research primarily focuses on a specific setup: a 2-finger antipodal gripper mounted on a robot arm, which is prevalent in industrial applications due to its robust design and simple working principle. This setup features only one in-hand degree of freedom (DoF), significantly limiting its kinematics compared to more complex multi-finger or multi-joint setups, e.g.~\cite{koenig2022role,pavlichenko2023deep,pitz2023dextrous,garcia2019tactilegcn}. Consequently, achieving a form-closure grasp is nearly impossible in our setup. This limitation requires incorporating the motion of the robot arm into the grasping policy, rather than relying solely on in-hand manipulation. Moreover, depending on the specific geometries, the 2-finger antipodal grasping often produces a much sparser reward due to the fewer contact points. The small contact surface, however, enables sliding motion around the object's surface for exploration. This scenario imposes high demands on the integration of visual (for global geometry) and tactile sensing (for local adjustment) and on the efficiency of the learning algorithms. Therefore, this study uses off-policy model-free reinforcement learning to tackle grasping tasks.

The main contributions can be summarized as follows: 
\begin{itemize}
\item A parallelized simulation/training pipeline for 2-finger antipodal grasping tasks with tactile feedback. 
\item Comparison of grasping performance using different tactile sensing layouts, measured quantities, environmental noise, and learning architectures. 
\item A practical guide on using tactile sensing in 2-finger antipodal grasping tasks to enhance performance. 
\end{itemize}

\vspace{-2mm}
\section{Related Work}\label{sec:relwork}

\paragraph{Learning-based methods for robotic grasping}

Robotic grasping describes the entire process from perception via action execution to holding the target object stably in hand for further manipulation under certain constraints.
This problem is extensively studied, especially for industrial pick-and-place scenarios \cite{kleeberger2020survey,xie2023learning}. 
Large datasets have been created and annotated using real sensor data or simulated settings, enabling the supervised learning of static grasping configurations \cite{fang2020graspnet,dai2023graspnerf}. The learned models are accurate, efficient, and generalize well to novel objects based on vision input alone. However, their output is usually a single selected grasping pose, requiring additional path planning to execute the grasp. Reacting to perturbations or failures due to perception errors is challenging in this setup. Multiple deep reinforcement learning-based grasping methods have been proposed \cite{kalashnikov2018scalable,quillen2018deep,pavlichenko2023deep,koenig2022role} to overcome this problem using reactive policies. These methods mostly use
visual perception as sensing input to provide feedback for failure correction. In recent works, tactile sensing has been introduced to RL-based grasping pipelines to reduce the gap between visual perception and the state of the real environment. Sim-to-real manipulation tasks have utilized raw binary contacts for better transfer \cite{ding2021sim}. Binary tactile grids and proprioceptive information have been used for grasping adjustment on a 3-finger manipulation platform \cite{wu2020mat}. Vulin et al.\ use intrinsic rewards based on tactile sensing that helps exploration in RL settings \cite{vulin2021improved}. Fusing force-torque sensing and visual input using transformers has also been shown to improve grasping performance \cite{chen2022visuo}. However, the literature lacks a detailed comparison of the influence of different tactile modalities in the context of 2-finger reach-and-grasp tasks solved with reinforcement learning, which we present in this paper.

\paragraph{Tactile sensors for robotics}
Tactile sensors with different modalities have been developed for robotics, especially for providing force feedback \cite{li2020review}. Sensors like Minsight \cite{Andrussow23-AIS-Minsight}, Insight\cite{Insight}, Piacenza et al. \cite{piacenza2020disco} or DenseTact 2.0 \cite{wo2023densetact20} provide high-resolution all-around tactile perception with a sensing frequency up to 60 Hz. BioTac \cite{Lin2013EstimatingPO},  GelSight \cite{yuan2017gelsight}, and GelSlim \cite{taylor2022gelslim} provide
 tactile information on a flat or curved surface, which can be used to extract local textures, object orientations \cite{takahashi2024stable} and force information \cite{dong2021high}. Discretized sensor arrays, like USkin \cite{funabashi2020stable} and binary taxel arrays \cite{yang2023tacgnn}, are used to provide low-resolution tactile information. Force-torque sensors are also used for simple force feedback, e.g.\ in \cite{chen2022visuo} and \cite{pitz2023dextrous}. A comparison across sensing area and measured quantities can be found in \cref{fig:tactile-sensor-classification}. In this paper, we create a prototypical simulation to abstract away the specific physical properties of tactile sensors to focus on their high-level differences in sensing area, sensing resolution and representation of the measured forces. For our real-robot experiment, we use the vision-based tactile sensor Minsight~\cite{Andrussow23-AIS-Minsight}, as a representative tool that can deliver different modes of force-based tactile information.

\paragraph{Evaluation of tactile sensing for RL-based grasping}
Multiple force stability criteria are compared by Koenig et al.~\cite{koenig2022role} for multi-finger re-grasping setups using PPO \cite{schulman2017proximal}. However, only a single force per finger link has been presented with 3 data representations. Clustered and full tactile sensing for in-hand manipulation tasks are compared by Melnik et al.\cite{melnik2019tactile} on the Shadow Dexterous Hand. Three visual-tactile encoders are compared by Chen et al. \cite{chen2022visuo} for grasping and manipulation tasks in a model-based RL setup with force-torque-based tactile sensing. The existing works mostly focus on multi-finger setups and do not provide a systematic comparison of tactile sensing areas and sensed quantities. 
Our work closes this gap by conducting well-designed experiments to assess the importance of the sensing area, the quantities detected, the learning architectures and the memory length for robust and generalizable grasping.


\section{Approach}\label{sec:approach}
To study which tactile information is beneficial for learning robust antipodal grasping behavior, we designed a Reinforcement Learning (RL) setup, a training pipeline, and a simulation that allows selection and approximation of tactile sensing.

\subsection{Modeling antipodal grasping as an RL problem}
We consider a robot arm with $n_{\text{arm}}$ degrees of freedom (DoF) equipped with a 2-finger antipodal gripper with sensor inputs such as proprioceptive information, monocular camera image and tactile information.
The aim is to find a policy $\pi$ that performs reaching and robust grasping of an object, selected from a larger object family.
One advantage of learning such a policy is that it can learn to react to perception errors and generally deal with uncertainties. 
There are two main challenges in creating this RL setup. We need a good simulation for massive data collection and we need to define a reward function that allows learning the desired task. 
We propose to perform a stability test as part of the reward computation
such that stable grasps can be obtained.

We follow the standard RL framework using a Markov Decision Process (MDP) given by $\{S,A,P,R,\gamma\}$ to describe the environment with the state space $S \in \mathbb{R}^{n_s}$, the action space $A \in \mathbb{R}^{n_{\text{arm}}+1}$, the transition function $P$ implemented by the simulation, the reward function $R: A \times S \rightarrow \mathbb{R}$ and the discount factor $\gamma$. 
We optimize the action policy $\pi: S \rightarrow A$ to maximize the discounted future reward. 
The task is formulated as an episodic problem, \ie the task must be completed within a fixed duration of the episode $T$. 

\textbf{The state} contains four parts, $s = [s_{\text{pp}}, s_{\text{visual}},s_{\text{tactile}}, s_{\text{step}}]$ where $s_{\text{pp}}$ is the robot proprioception data, which includes the tool center point (TCP) pose in Cartesian space and the opening size of the gripper. 
The visual state $s_{\text{visual}}$ contains the object type encoding and object pose in Cartesian space. 
The tactile state $s_{\text{tactile}}$ contains tactile information depending on the experimental setup shown in \cref{tab:ex-configuration}. 
The time information $s_{\text{step}}$ is a unified scalar proportional to the remaining episode steps.

\textbf{The action} consists of all joint positions and the width of the gripper opening $a_t = [j_1,...,j_{n_{\text{arm}}},j_{\text{gripper}}]$. 
Other works, e.g.\ \cite{wu2022grasparl}, consider operational space control, which requires another low-level motion controller that is unaware of the current task potentially jeopardizing reactivity.

\textbf{The reward} contains two terms: $r_{\text{grasp}}$ and $r_{\text{aux}}$ with $r_{\text{grasp}} \gg r_{\text{aux}}$.  
The main reward $r_{\text{grasp}}$ captures task success and is only provided at time step $T$.
Its computation requires a stability-checking process that we detail below.

The auxiliary reward $r_{\text{aux}}$ is available at each step and provides reward shaping to achieve faster convergence. It contains terms for approaching, touching, and applying gripping forces and penalty as detailed in \cref{tab:reward-definition}.

\begin{table}[H]
\centering
\vspace{-.5em}
\caption{\uppercase{Definition of auxiliary rewards.}} \label{tab:reward-definition}\vspace{.2em}
\adjustbox{max width=\linewidth}{
\begin{tabular}{@{}c|c}
\toprule
Name & Weight $\cdot$ Reward \\
\midrule
touch & $10\cdot(0 \text{ if both fingers in contact} \text{ else} -1)$ \\
approach & $10\cdot ((d^l_t + d^r_t) - (d^l_{t-1} + d^r_{t-1}))$ \\
force & $1\cdot d_\theta(f^l_t, f^r_t) - d_\theta(f^l_{t-1} + f^r_{t-1}))$ \\
\multirow{2}{*}{penalty} & $-1\cdot g(\frac{1}{f_{\text{max}}}\mathrm{clip}(\max(f^l_t, f^r_t)-f_{\text{penalty}}, 0, f_{\text{max}})/g(1)$,\\
& where $g(x)=1-\exp(-x)$, $f_{\text{max}} = 3 \cdot f_{\text{penalty}}$\\
\bottomrule
\end{tabular}
}
\vspace{-2mm}
\end{table}

\textbf{The stability check} allows us to assess whether a stable grasp was obtained. Once the learned policy has executed the grasp, we perform an action sequence that lifts the arm and holds it in place without changing the gripper orientation, applying random forces to the object and then waits for a certain time. During the entire process, we monitor the contact of both fingers with the object and accumulate the time $t_{\text{inhand}}$, during which the object remains in the grasp. The reward is computed as $r_{\text{grasp}} = 1000\cdot \frac{t_{\text{inhand}}}{t_{\text{total}}}$, where $t_{\text{total}}$ is the duration of the stability check.

\textbf{Demo injection} means that the experience buffer is pre-filled with trajectories with a precomputed grasp position with a similar method to \cite{fang2020graspnet}. This speeds up training but does not change the asymptotic performance.

\begin{figure}
\begin{subfigure}{.58\linewidth}
  \caption{Isaac Simulation} 
  \includegraphics[width=1\textwidth]{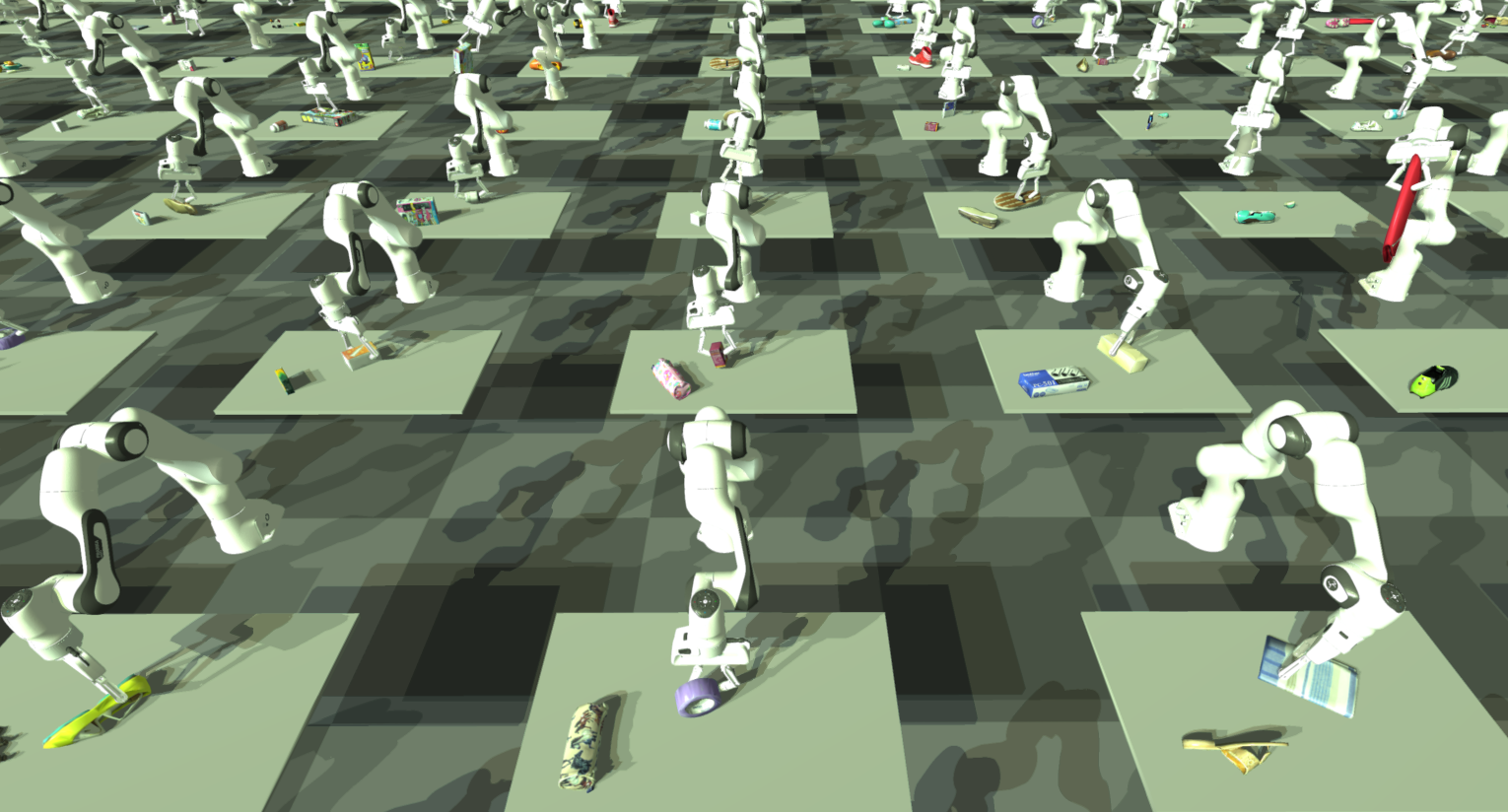}
  \vspace{0.05\linewidth}
  \label{fig:sim-table-top}
\end{subfigure}%
\hspace{0.01\linewidth}
\begin{subfigure}{.40\linewidth}
 \caption{Objects}
  \includegraphics[width=1\textwidth]{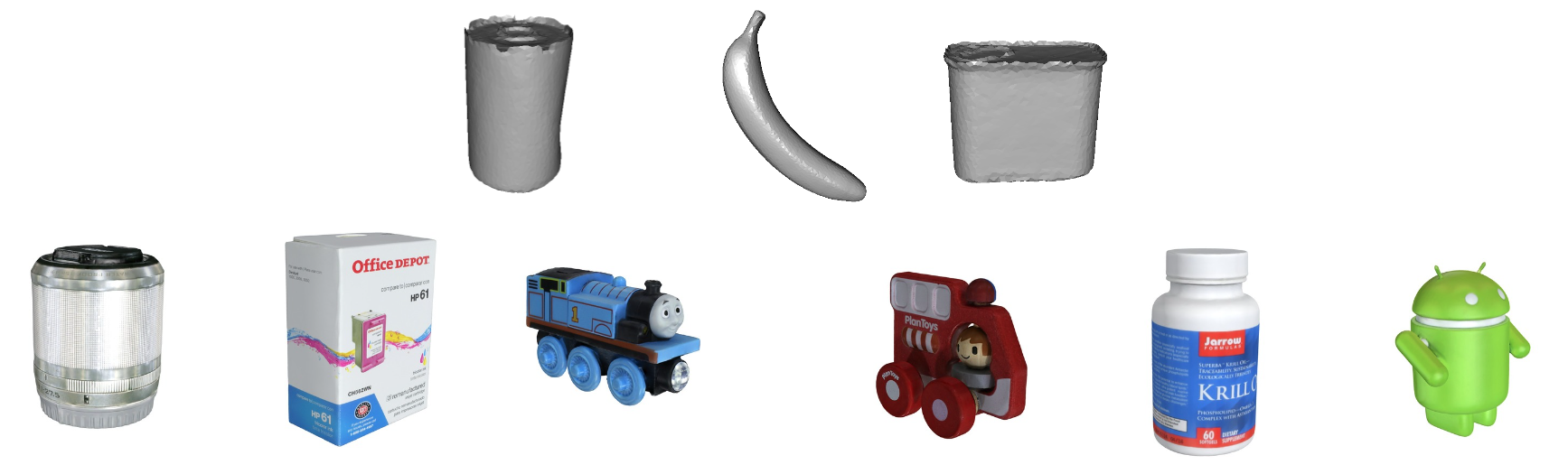}
  \label{fig:objects}
  \vspace*{-1em}
  \caption{Sensor Approximation}  
  \includegraphics[width=1\textwidth]{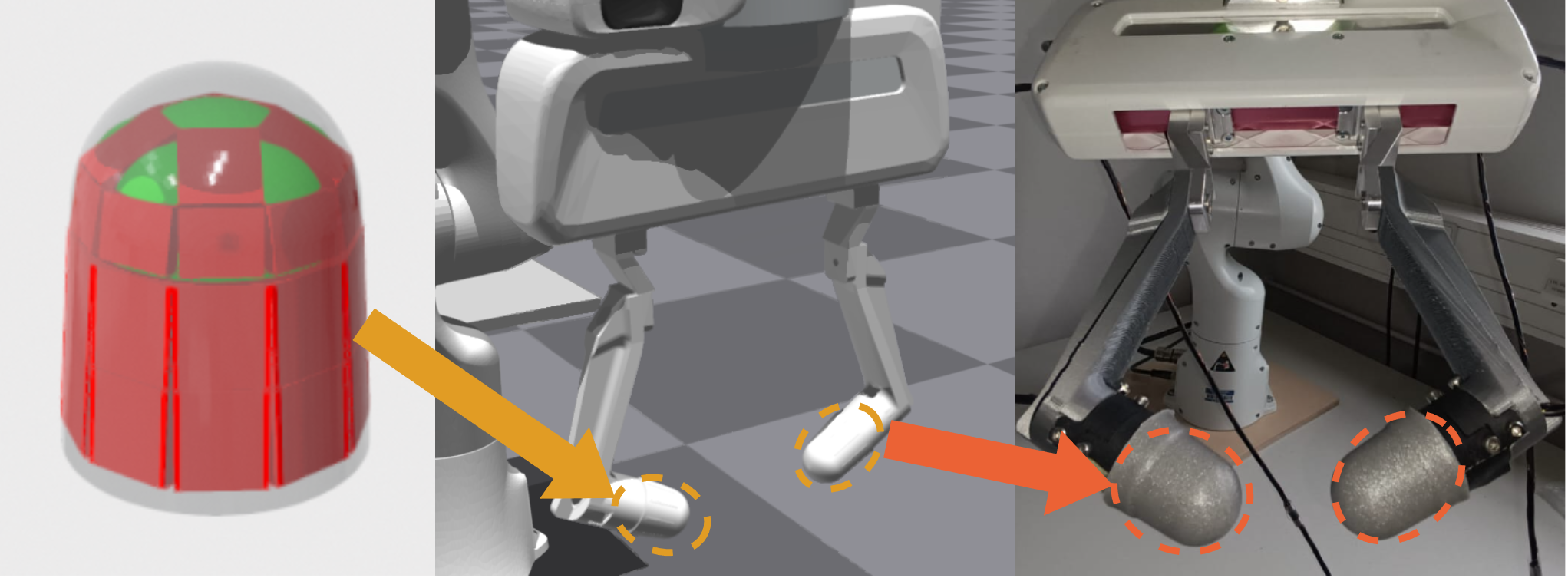}
  \label{fig:approx-tactile-sensor}
\end{subfigure}%
\vspace*{-1em}
  \caption{Simulation setup: (a) environment for tabletop grasping with panda robot and antipodal gripper.
  (b) Top row: three main objects; Bottom row: samples from 100 objects.
  (c) Approximation of the tactile sensor next to the real Minsight sensor \cite{Andrussow23-AIS-Minsight}.
  }
\vspace*{-2em}
  \label{fig:simulation}
\end{figure}

\subsection{Simulation Environment}
Our experiments are conducted within a simulated framework using Isaacgym \cite{makoviychuk2021isaac}.
We create multiple independent environments, each with a Franka Panda robot ($n_{\text{arm}}=7$) placed in the middle of the area. A 2-finger antipodal gripper, equipped with a tactile sensor on each finger, is mounted on the last link of the robot. A camera mounted on the wrist delivers RGB images. 

As shown in \cref{fig:sim-table-top}, the grasping task is a tabletop grasping scenario. The objects are taken from the YCB \cite{singh2014bigbird} and Google scanned dataset \cite{downs2022google}. 
Following the demo-injection method, stable placing poses and robust grasping poses (used for computing good initializations) are pre-calculated based on the homogeneous mass distribution assumption. Target objects are sampled from a selected subset of the object dataset based on graspability. Rigid-body collisions between the finger and objects are assumed. During training, the target object is placed in a randomly sampled pre-calculated stable orientation and random position within a reachable range. The policy runs at 20\,Hz and raw visual and tactile information is collected from the simulation at each step.

\subsection{Sensor Approximation}\label{sec:SensorApproximation}
\begin{figure}
\centering
\begin{minipage}[t]{.95\linewidth}
  \centering \ \\[-1em]%
  \includegraphics[width=\linewidth]{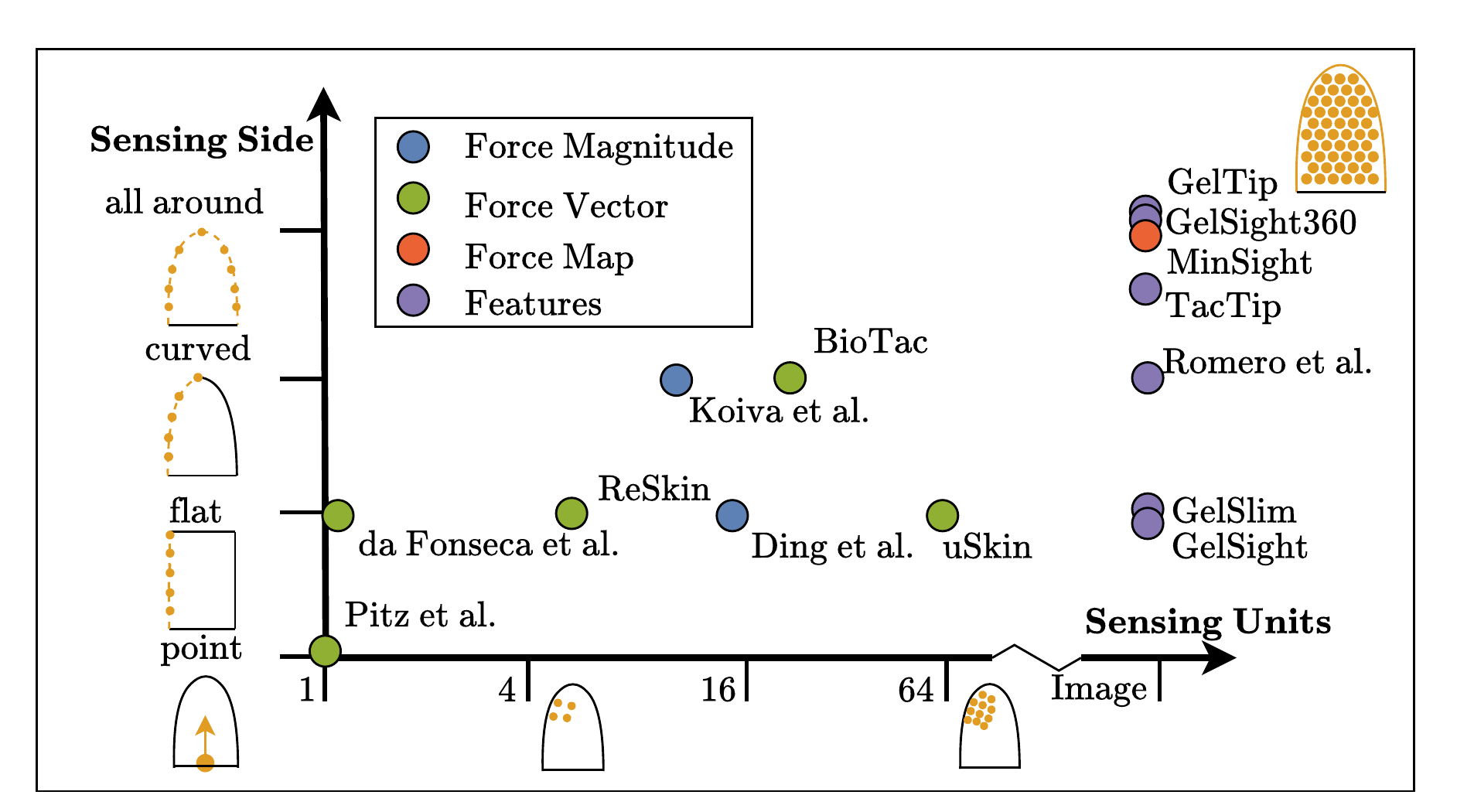}
  \caption{Tactile sensor classification according to the sensing area (side/coverage and units/resolution) and sensing type (color). Sensors created or used by:
Romero et al.\cite{romero2020soft},
TacTip~\cite{ward2018tactip},
ReSkin~\cite{bhirangi2021reskin}, 
da Fonseca et al. \cite{da2022tactile}, 
Ding et al. \cite{ding2021sim},
Pitz et al. \cite{pitz2023dextrous},
Koiva et al. \cite{koiva2013highly},
uSkin~\cite{funabashi2020stable},
BioTac~\cite{Lin2013EstimatingPO},
GelSight~\cite{yuan2017gelsight},
GelSlim~\cite{taylor2022gelslim},
GelTip~\cite{gomes2020geltip},
GelSight360~\cite{tippur2023gelsight360},
Minsight~\cite{Andrussow23-AIS-Minsight}.
}
  \label{fig:tactile-sensor-classification}
\end{minipage}%
\end{figure}

To investigate which kind of tactile information is most beneficial for learning to grasp, 
we first categorize existing sensors by sensing area and measured tactile quantity (see \cref{fig:tactile-sensor-classification}).   
The result shows that the number of sensing points on the surface and the type of data provided by the sensor differ vastly.
For our experiments, we select the Minsight tactile sensor \cite{Andrussow23-AIS-Minsight} as it features a large sensing area of $1740\,\text{mm}^2$ with all-around sensing of normal and shear forces and a high sampling rate of $60$\,Hz. We believe it is highly suitable for simulating other sensors with smaller sensing areas or less complex sensing data.  
For the simulation, we need to approximate the sensor's shape and working principle. 
Finite element simulations in Isaacgym are too slow for large data collection, and modeling the surface as a mesh yields very unstable mesh-mesh interactions when grasping objects.
Thus, we resort to approximating the shape with primitive geometries, as shown in \cref{fig:approx-tactile-sensor}. The model consists of 45 cuboids and 5 spheres. Each of these primitives can sense a net force vector independently, which gives a total resolution of 50 force vectors per sensor. 

\begin{figure}
\centering 
\begin{minipage}[t]{.98\linewidth}
  \includegraphics[width=\linewidth]{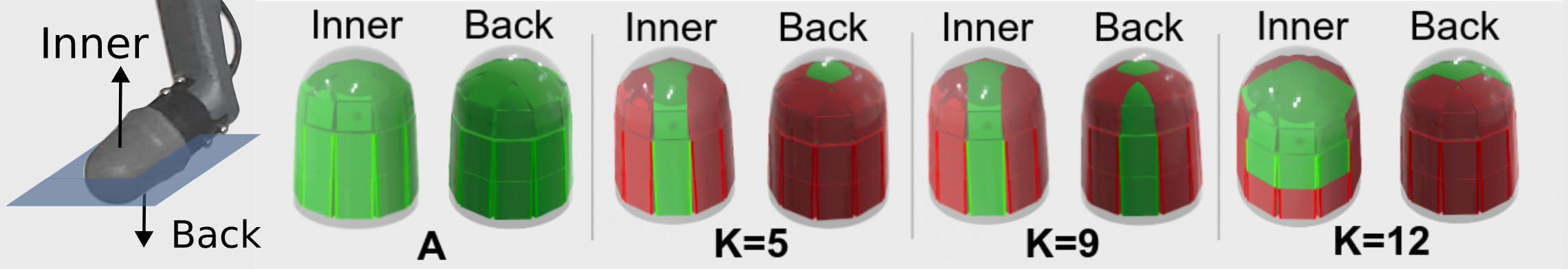}   
  \caption{Approximation of tactile sensors with different sensing areas, left (view from the inner side), right (view from the back side). A: the whole area of the sensor gives force feedback (used for detailed force map or sum of forces). 
  K=5: force feedback only for sites on inner center line. 
  K=9: extended center line to include also the back side. 
  K=12: provides dense feedback for the area most touched on the inner side of the sensor }
  \label{fig:sensor_type}
  \vspace*{-2em}
\end{minipage}
\end{figure}

This simulation approximates various force-based tactile sensors, regardless of the sensing principle.
We classify existing sensors  along three dimensions in \cref{fig:tactile-sensor-classification}: the number of sensing units/resolution, then sensing area and the type of information provided.
For simplicity we consider three categories for the sensing coverage: 
1)~\textbf{A single unit sensor} treats the fingertip as a single unit, sensing the total force. It mimics a force-torque sensor at the finger's base.
2)~\textbf{A K feature sensor} divides the fingertip into K regions. Depending on the density, different K values and taxel locations are chosen as shown in \cref{fig:sensor_type}. It mimics various tactile sensors with independent taxels (\cref{fig:tactile-sensor-classification}).
3)~\textbf{The full-range sensor} is sensitive across the whole sensor body and outputs a high-resolution force map.\looseness-1

For the single unit sensor and the K feature sensor, three different measurement quantities are extracted as sensor readings:
1)~\textbf{Binary signal} (denoted as \textbf{B}) returns 1 if the taxel is touched and 0 otherwise. 
Sensors used in \cite{gianoglio2023trade, lee2024dextouch} belong to this class. 
2)~\textbf{Magnitude signal} (denoted as \textbf{M}) returns the magnitude of the gross force being applied to each taxel. Sensors used in \cite{ding2021sim} \cite{koiva2013highly}  belong to this class.   
3)~\textbf{Force vector signal} (denoted as \textbf{V}) returns a force vector $[f_x,f_y,f_z]$ for each taxel. It contains the magnitude and orientation of the forces being applied. Sensors used in \cite{da2022tactile,Andrussow23-AIS-Minsight} belong to this class.\looseness-1

Abbreviations are summarized in \cref{tab:tactile-state}.


\section{Experiments and Analysis}\label{sec:ExperimentsSetup}

We designed a set of experiments, listed in \cref{tab:ex-configuration}, that allow us to study how tactile feedback impacts the learning of stable grasping.  
It was shown that grasping can be achieved solely with visual perception data~\cite{dai2023graspnerf,fang2020graspnet}.
So our first experiment in \cref{sec:ex1} investigates whether tactile sensing can still help in the case of perfect visual perception. 
 
Grasping with visual-only perception data is mostly vulnerable to visual occlusion and perception imperfection. In addition, it has a high computational demand and often introduces a time-lag. 
Thus, in the second experiment in \cref{sec:ex2} we investigate how different tactile sensing compensates for visual perception imperfection. 

The tactile sensor's resolution and mounting position provide different contact perspectives. 
The all-around sensor covers a larger area but comes with higher complexity and computational cost compared to the point-wise sensor. The third in \cref{sec:ex3} determines which design offers the best learning performance.

RL methods usually require a large amount of data for generalization across tasks. 
Without specific processing (e.g.\ object detection, pose estimation, etc.), visual perception usually has a higher variance across domains.
Tactile sensing, meanwhile, can provide more consistent representations. 
Whether this feature leads to a better generalization to different objects is investigated in the fourth experiment in \cref{sec:ex4}.

\begin{table*}[t]
\parbox{.61\linewidth}{
\centering
\caption{\uppercase{Experiment configuration} } \label{tab:ex-configuration}\vspace{.2em}
\adjustbox{max width=\linewidth}{
\begin{tabular}{@{\hspace{1pt}}c@{\hspace{1pt}}|cc@{}c@{}c@{}c@{\hspace{2pt}}c@{}}
\toprule
 ID  & Visual State& Tactile Sensor Type& K & Visual Noise & Memory Length & Objects \\
\midrule
1 & $[p_{\text{obj}},q_{\text{obj}}, \mathbf{1}_{\text{obj}}]$ & B,V,BK,VK,E & 5 & -- & 5 & 3 \\
2 & $[p_{\text{obj}},q_{\text{obj}}, \mathbf{1}_{\text{obj}}]$& B,M,V,BK,MK,VK,E & 5 & OU, offset & 1,5,10 & 3 \\
3 & $[p_{\text{obj}},q_{\text{obj}}, \mathbf{1}_{\text{obj}}]$& M,V,MK,VK,E & $5,9,12$ & OU, offset & 5 & 3 \\
4 & $[\mathrm{Img}_{\text{wrist}}]$& V & -- & -- & 5 & 10, 100 \\
5 & $[p_{\text{obj}},q_{\text{obj}}, \mathbf{1}_{\text{obj}}]$& B,M,V,BK,MK,E & 5 & OU, offset/env noise & 5 & 3 \\
\bottomrule
\end{tabular}
}
}
\hfill
\parbox{.37\linewidth}{
\centering
\vspace{-.5em}
\caption{\uppercase{The tactile state}} \label{tab:tactile-state}\vspace{.2em}
\adjustbox{max width=\linewidth}{
\begin{tabular}{@{}c|lc@{}}
\toprule
Type & Description & Size  \\
\midrule
 B & overall binary touch signal (touch / no touch) & 1*2 \\
 M & overall force magnitude
& 1*2 \\
 V & overall 3D force vector 
 & 3*2 \\
BK & binary touch signal for each feature point 
 & K*2\\
MK & force magnitude for each feature point
 & K*2\\
VK & 3D force vector for each feature point
 & 3K*2 \\
\bottomrule
\end{tabular}
}
\vspace{-2mm}
}
\vspace{-2mm}
\end{table*}

\begin{table}[H]
\centering
\vspace{-.5em}
\caption{\uppercase{Parameters for Deep RL methods.}} \label{tab:training-params}
\vspace{-.3em}
\adjustbox{max width=\linewidth}{
\begin{tabular}{c|ccccc}
\toprule
Method & Learning Rates & Model & Hidden Layer Sizes & Buffer Length\\
\midrule
 SAC & $1e^{-3}$/$1e^{-4}$ & MLP & [256, 256, 512] & 2M \\
\midrule
 MPO & $1e^{-3}/1e^{-3}/1e^{-2}$ & MLP & [256, 256, 512] & 2M \\
\bottomrule
\end{tabular}
}
\vspace{-2mm}
\end{table}

It is also important to know if the conclusions depend on the particular RL algorithm used. 
For this reason, we run our experiments both with SAC \cite{haarnoja2018soft} and MPO \cite{abdolmaleki2018maximum}. While SAC is a classic off-policy RL method, MPO is proven to be highly efficient for large dimensional state space \cite{schumacher2022dep, mankowitz2019robust}. Important training parameters are listed in \cref{tab:training-params}. All experiments are tested across multiple seeds (4-8 depending on time consumption). To make the conclusion stronger, we conduct sim-2-real experiments on real hardware in \cref{sec:ex5}.

\subsection{Is tactile sensing beneficial with perfect visual perception?}\label{sec:ex1}

\begin{figure}[h]
\begin{minipage}[t]{.64\linewidth}
  \ \\[-1em]
  \centering
  \includegraphics[width=\linewidth]{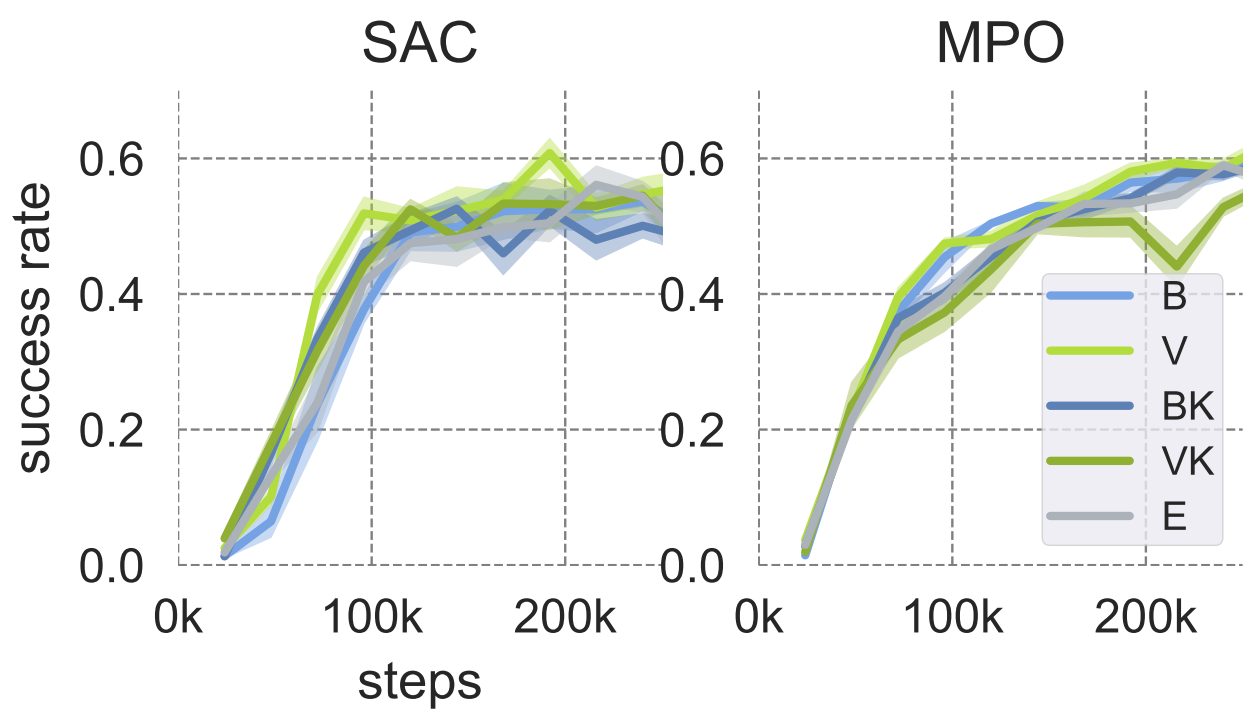}
  \caption{Training with ideal visual and tactile sensing for SAC and MPO.}
  \label{fig:ex1}
\end{minipage}
\begin{minipage}[t]{.33\linewidth}
  \ \\[-1em]
  \centering
  \includegraphics[width=\linewidth]{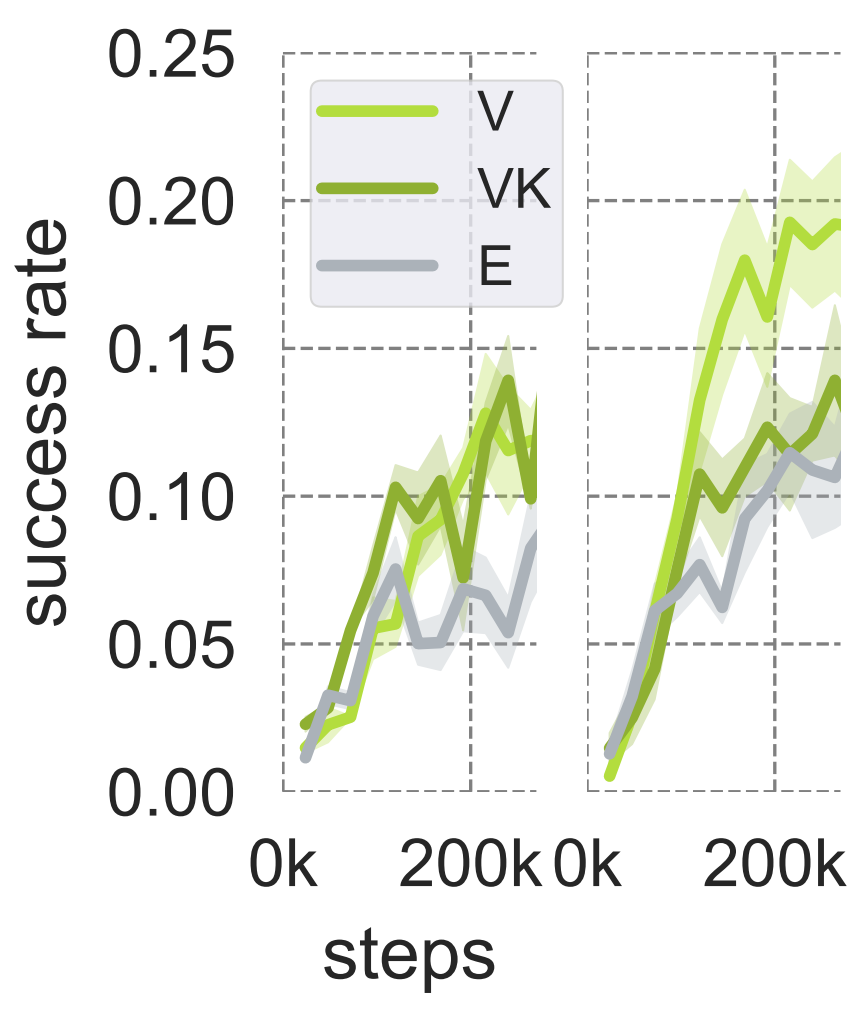}
  \caption{Comparison between 2 memory lengths using SAC:left:1, right:5.}
  \label{fig:ex2_memory_length}
\end{minipage}
\vspace{-.5em}
\end{figure}

In this experiment, the most condensed, but perfect representations of visual perception are chosen: the accurate pose of the target object and the object type as a one-hot encoding. 
Three objects are chosen for the grasping tasks. 
They contain enough surface details for exploration while keeping the geometric shape simple, see \cref{fig:objects}. 
As a first step, we checked the success rate of the precomputed grasps.
Given our simulation and our aggressive stability checking, the average success rate is only around 15\%, which warrants the need for RL. 
The state contains five successive time step observations $s_{\text{visual}} = [o_{t-4}, o_{t-3},..., o_t]$ to keep a memory of the environment. With demo-injection, the policy can converge in 300k training steps using a fixed learning rate and achieve about 60\% success rate.

\textbf{Analysis:}
As shown in \cref{fig:ex1}, policies using different representations of tactile information reach a similar success rate with ideal visual perception. 
This indicates that tactile sensing does not improve in-distribution performance when vision is perfect. We also see that SAC is saturating earlier and has a higher variance than MPO.
Note that the tactile information for $\mathrm{VK}$ occupies $61.2\%$ of the state-space, but that does not seem to have a strong impact on performance. 

\subsection{Does tactile sensing compensate for imperfect vision?}\label{sec:ex2}

In reality, visual perception is never perfect as it suffers from occlusions, suboptimal lighting conditions, shadows, etc. 
In this experiment, we simulate this by considering two types of noise. 

\begin{figure}
\centering
\vspace*{-2em}
\begin{subfigure}{.22\linewidth}
  \caption{0.3 OU} 
  \includegraphics[width=1\textwidth]{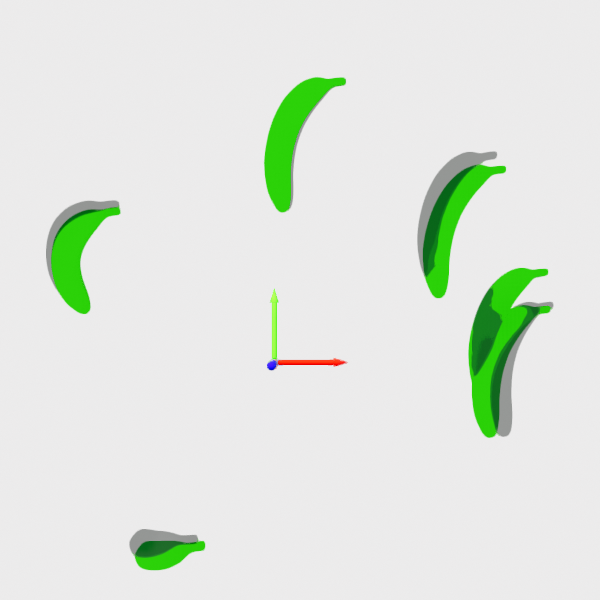}
  \label{fig:supp-noise1}
\end{subfigure}
\vspace{0.01\textwidth}
\begin{subfigure}{.22\linewidth}
  \caption{0.6 OU} 
  \includegraphics[width=1\textwidth]{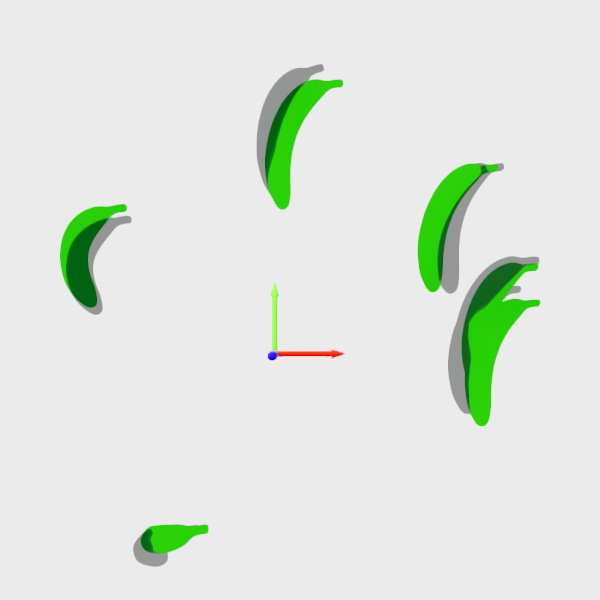}
  \label{fig:supp-noise2}
\end{subfigure}
\vspace{0.01\textwidth}
\begin{subfigure}{.22\linewidth}
  \caption{0.3 Offset} 
  \includegraphics[width=1\textwidth]{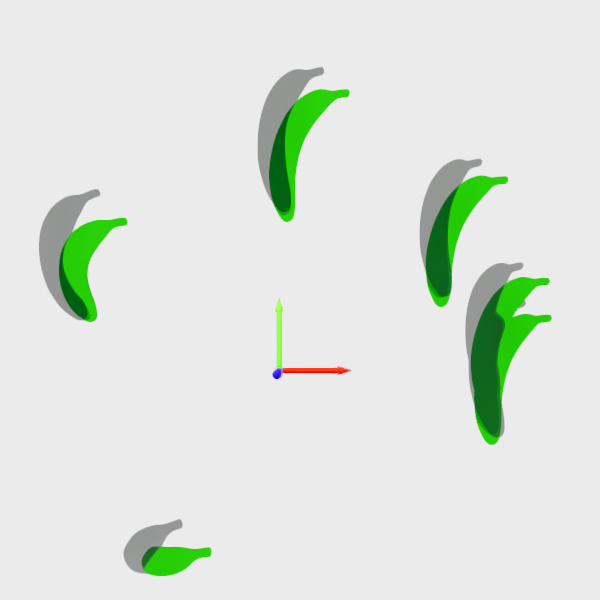}
  \label{fig:supp-noise3}
\end{subfigure}
\begin{subfigure}{.22\linewidth}
  \caption{0.3 OU + Offset} 
  \includegraphics[width=1\textwidth]{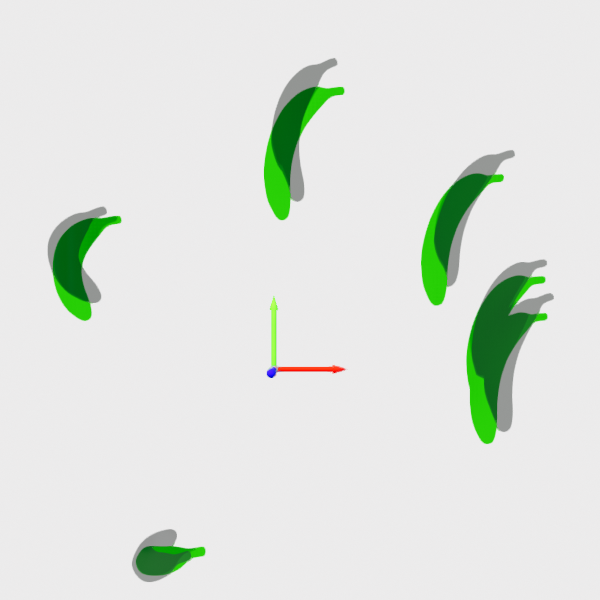}
  \label{fig:supp-noise4}
\end{subfigure}
\vspace*{-2em}
  \caption{Sampled noises of different magnitudes and combinations. Green objects show the ground truth poses. Grey objects show noised poses.}
  \label{fig:noise}
\vspace*{-2em}
\end{figure}

An offset noise (\cref{fig:supp-noise4}) is applied for each environment to simulate the potential sensor calibration error: $X_t = X_t + \theta_{\text{off}} X_\delta$, where $X_t = [p_{t, \text{obj}},q_{t, \text{obj}}]$ (position and orientation), $X_\delta$ is randomly chosen at the beginning of each episode and remains the same for the rest of the episode, $\theta_{\text{off}}$ is a hyperparameter that tunes the magnitude of the offset noise. 
The second noise follows an Ornstein-Uhlenbeck (OU) process and is applied to each step to simulate the systematic noise brought by the vision system: $dX_t = \theta_{\text{ou}}(\mu_{t} - X_t)dt + \sigma_{\text{ou}} dW_t$ where $\mu_t$ is the ground truth pose to reach, $dW_t$ is sampled from a Gaussian distribution, $\theta_{\text{ou}}$ and $\sigma_{\text{ou}}$ are hyperparameters that tune the magnitude and tracking behavior introduced by the OU noise.
We hypothesize that, with stronger noise being applied, the policy should increasingly rely on the more robust tactile signal.

First, we want to make sure that simply using memory cannot compensate for the effects of the noise. 
We compare a single frame input $s_{\text{vision}} = [o_t]$ with  $n$-successive-frames as described in \cref{sec:ex1}, where $n \in [5,10]$.

\textbf{Analysis:}
As shown in \cref{fig:ex2_memory_length}, memory cannot compensate for the noise and 5 successive frames as memory provides the best performance. In an exploratory experiment with length equal to 10 shows longer memory leads to a decline in performance. 
This indicates that only short-term memory is necessary for grasping tasks which can also be observed in human behaviors \cite{scheidt2001learning}. 

\begin{figure}
  \centering
  \vspace{-1em}
  \includegraphics[width=0.9\linewidth]{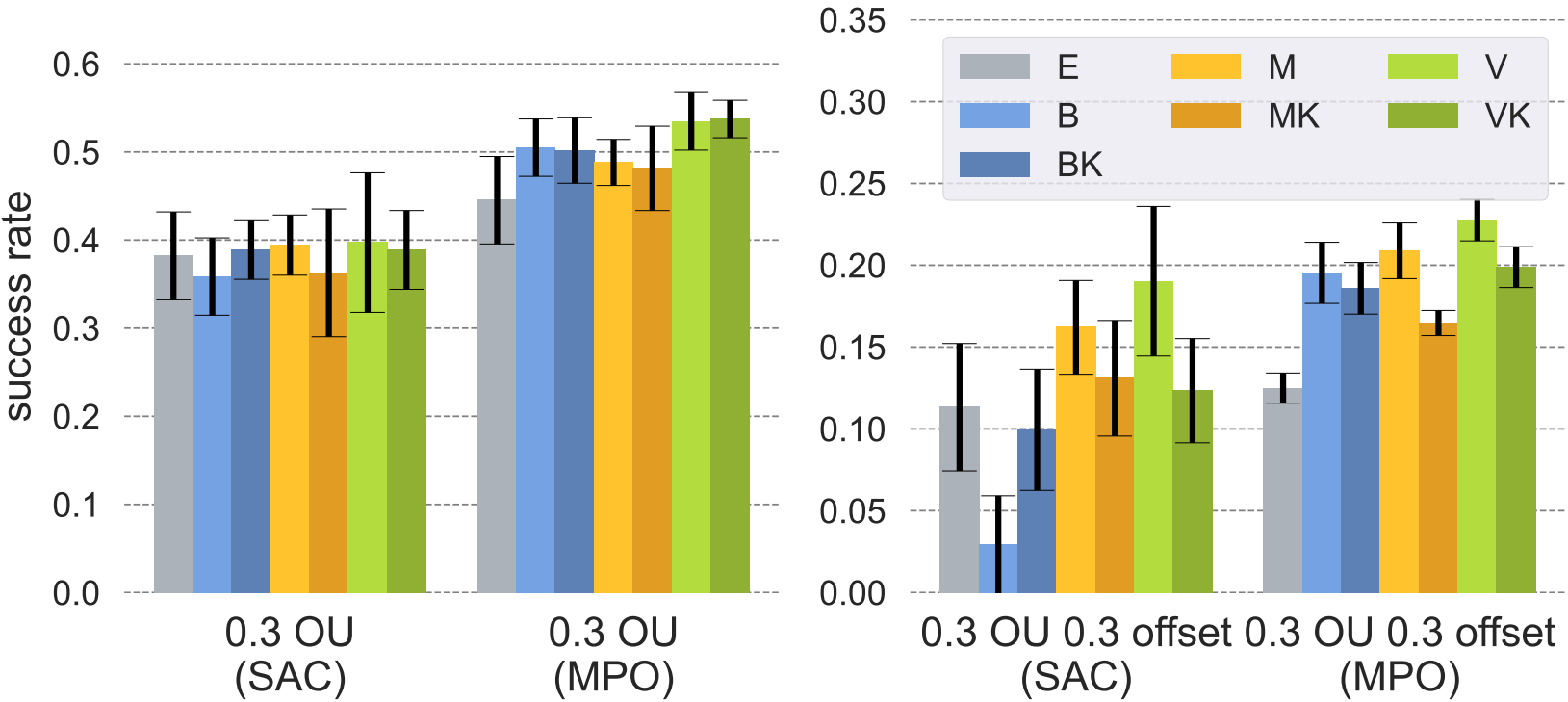}
  \caption{Comparison of different tactile information when visual perception is impaired by OU noise and offset noise.
  In both noise cases, tactile sensation leads to improved performance, in particular using MPO. SAC is sometimes brittle, in particular with the offset noise.
  }
  \label{fig:ex2}
  \vspace{-2em}
\end{figure}
As shown in \cref{fig:ex2}, pure OU noise affects the success rate of all setups. The policy can learn to use memory to reduce the uncertainty introduced by the OU noise. Compared to no tactile \textbf{E}, there is either no observable difference (SAC), or a small improvement (MPO) for the case with tactile sensing, indicating that the policy still relies mostly on visual perception under OU noise.

When offset noise is present, the tactile input enables a better low-variance performance, although generally with a lower success rate. In particular, MPO is able to handle the noisy perception well.
Especially, \textbf{V} and \textbf{VK} consistently outperform other tactile sensing types in all comparisons. 
This may be caused by the fact that force closure is often the only viable option in antipodal grasping scenarios. The grasping stability can be solely determined by examining the sum force exerted on each finger when the finger is only in contact with the object. While binary tactile sensing effectively registers contact, the ambiguity between internal and external touch necessitates the deployment of robust learning algorithms, as shown by SAC and MPO on the right side of \cref{fig:ex2}.

\subsection{Is a large sensing area necessary for 2-finger grasping?}\label{sec:ex3}
In this experiment, sensors with different cover areas are compared under the noisy vision condition (OU + offset). As shown in \cref{fig:sensor_type}, $K\in [5,9,12]$ are selected based on the classification of popular sensors. 
A description of the sensor configuration is given in \cref{sec:SensorApproximation}. 
To check the influence of the data complexity, experiments with the extra sum of the force ``\textbf{VK}+\textbf{V}'' are also conducted for each case. 

\begin{figure}
\centering
  \includegraphics[width=0.75\linewidth]{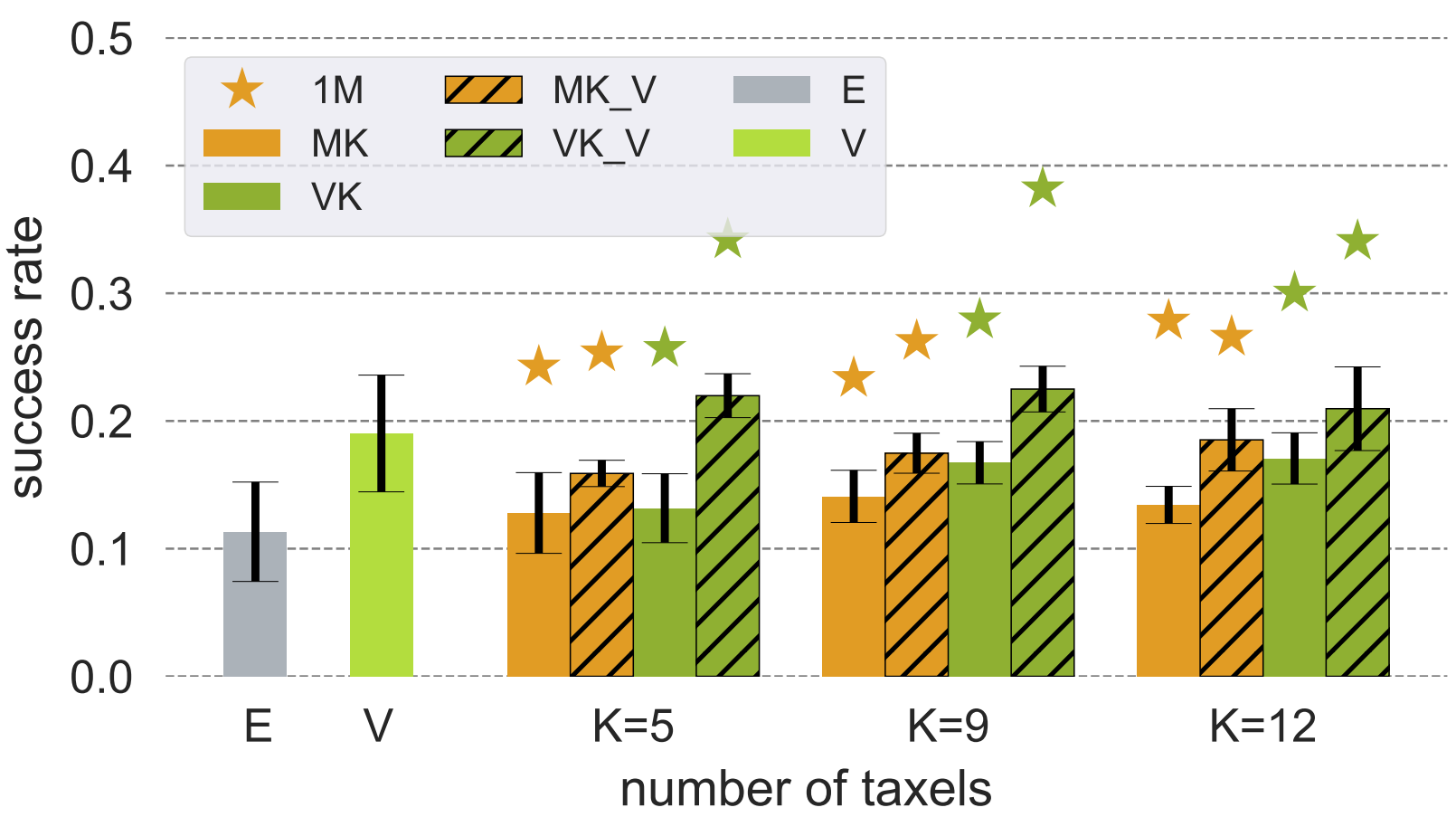}
  \vspace{-0.5em}
  \caption{Comparison of tactile sensing area using SAC and noisy visual information. Striped bars: taxel input + overall force vector ($\mathbf{V}$).
  }
  \vspace{0.5em}
  \label{fig:ex3_tactile_area_compare}
\end{figure}

\textbf{Analysis:}
\Cref{fig:ex3_tactile_area_compare} presents the results and suggests three conclusions: 
1)~Fewer taxels are worse: $\mathbf{VK}^{12} \sim \mathbf{VK}^{9} \ge \mathbf{VK}^5$; 
2)~Additional overall force helps: $\{\mathbf{VK}^5,\mathbf{VK}^9,\mathbf{VK}^{12}\}+\mathbf{V}\ge\mathbf{V}>\mathbf{VK}^5,\mathbf{VK}^9,\mathbf{VK}^{12}$; 
3)~Force magnitude feedback does not help: $\mathbf{MK}^5 \sim \mathbf{MK}^9 \sim \mathbf{MK}^{12} \sim \mathbf{E}$ while $
\mathbf{E} < \{\mathbf{MK}^5,\mathbf{MK}^9,\mathbf{MK}^{12}\}+\mathbf{V} < \mathbf{V}$. 
Conclusion 1 shows that compared to simple local tactile feedback, either global (both sides) or denser local data achieve better performance. 
Conclusion 2 shows that with an extra concise global force, both cases achieve better performance than only with each separately. 
We hypothesize that the global force input discerns similar states with a different causal effect on grasping, making learning easier. Comparing conclusions 2 and 3, we can see that a good sensing quantity (e.g.\ \textbf{V}) is more important than having a large sensing area.

\begin{figure}
\hspace{0.015\textwidth}
\begin{minipage}[t]{.44\linewidth}
  \ \\[-1em]
  \centering
  \includegraphics[width=1\textwidth]{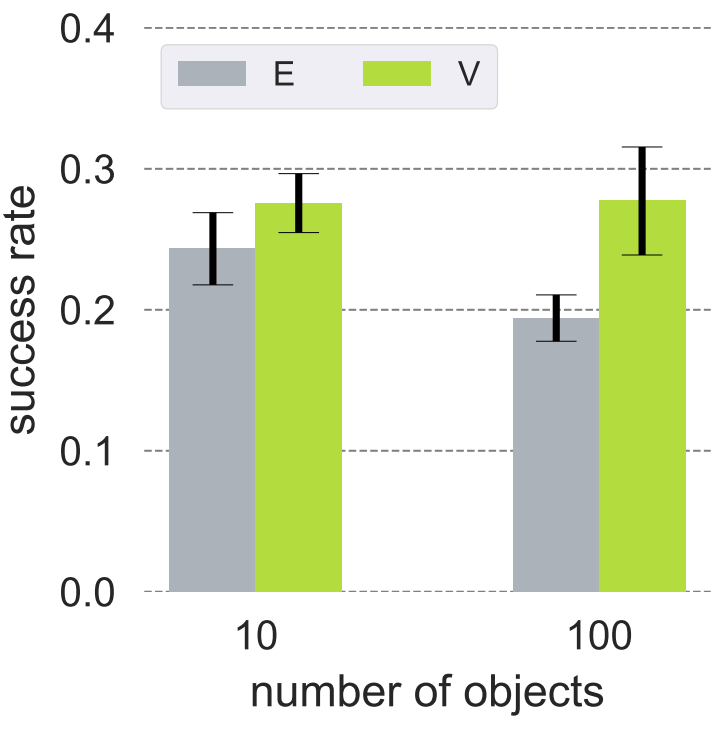}
  \caption{Comparison of generalization ability w/o tactile sensing using
  vision encoding (SAC).
  }
  \label{fig:ex4}
\end{minipage}
\begin{minipage}[t]{.44\linewidth}
  \ \\[-1em]
  \centering
  \includegraphics[width=\linewidth]{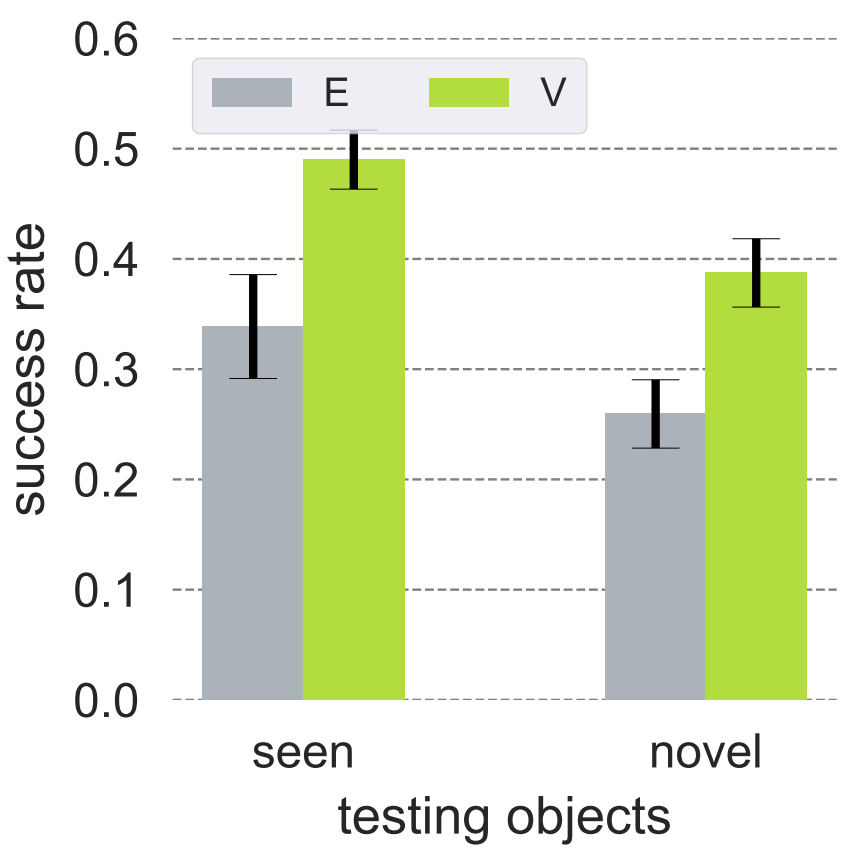}
  \caption{Generalization performance trained with 100 objects using a visual encoder and MPO.}
  \label{fig:ex4_bar_2}
\end{minipage}
\vspace{-2em}
\end{figure}

\begin{figure*}[tb]
\begin{minipage}[t]{.36\linewidth}
  \centering
  \ \\[-1em]
  \includegraphics[width=\linewidth]{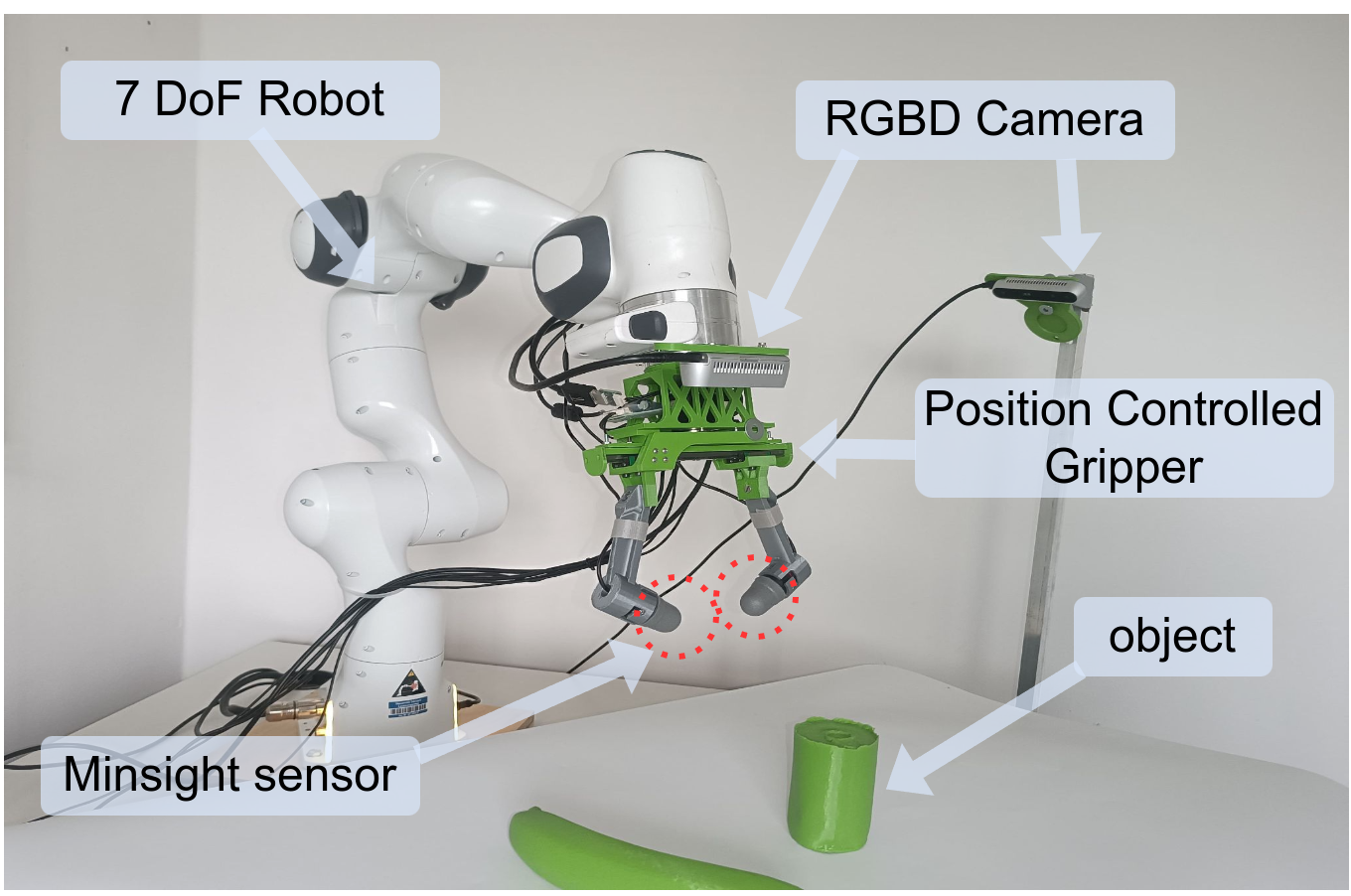}
  \caption{Real hardware setup.}
  \label{fig:real_ex_setup}
\end{minipage}%
\hspace{.02\linewidth}%
\begin{minipage}[t]{.61\linewidth}
  \centering
\captionof{table}{\uppercase{Average success rate for sim-2-real picking test. \\($\cdot$) success rates when the gripper is closed in a hard-coded manner after the policy rollout.}} \label{tab:ex-5}\vspace{.2em}
\adjustbox{max width=\linewidth}{
\begin{tabular}{@{}c|c|l|l|c|l|l|c|l|l@{}}
\toprule
\multirow{3}{*}{Tactile} & \multicolumn{9}{c}{Success Rate} \\ \cmidrule{2-10}
                              & \multicolumn{3}{c|}{Cola Can} & \multicolumn{3}{c|}{Box} & \multicolumn{3}{c}{Banana}\\
                              & sim & w/o disturb & w/ disturb & sim & w/o disturb & w/ disturb & sim & w/o disturb & w/ disturb\\
\midrule
 E  & 0.18 & 0.00 (0.30) & 0.00 (0.25) & 0.46 & 0.00 (0.35) & 0.00 (0.35) & 0.26 & 0.00 (0.45) & 0.00 (0.50) \\
 B  & 0.33 & 0.35 (0.55) & 0.25 (0.55) & 0.51 & 0.35 (0.55) & 0.20 (0.40) & 0.30 & 0.15 (0.55) & 0.30 (0.50)\\
 M  & 0.42 & 0.45 & 0.45 & 0.66 & \textbf{0.45} & \textbf{0.50} & 0.39 & 0.65 & \textbf{0.55}\\
 V  & \textbf{0.53} & \textbf{0.65} & \textbf{0.55} & \textbf{0.68} & \textbf{0.45} & 0.40 & 0.35 & \textbf{0.70} & 0.50\\
 BK & 0.45 & 0.20 (0.30) & 0.25 (0.40) & 0.53 & 0.20 (0.40) & 0.40 (0.45) & \textbf{0.46} & 0.15 (0.40) & 0.30 (0.45) \\
 MK & 0.41 & 0.05 (0.35) & 0.00 (0.50) & 0.63 & 0.25 (0.35) & 0.40 (0.45) & 0.34 & 0.20 (0.55) & 0.15 (0.55)\\
\bottomrule
\end{tabular}
}
\vspace{-2mm}
  
\end{minipage}
\vspace{-2em}
\end{figure*}
\subsection{Does tactile sensing enhance generalization?}\label{sec:ex4}
To answer this question, we attempt to grasp up to 100 objects.
Based on the setup of \cref{sec:ex1}, we change the visual representation to use low-resolution RGB images captured from the wrist camera to avoid the one-hot object encoding. 
We consider objects from the Google Scanned Objects~\cite{downs2022google} and  YCB~\cite{singh2014bigbird} and select the top 100 objects with a valid aspect ratio and the largest number of valid grasping poses calculated by a random grasping generation method.
An image encoder is pre-trained with data collected from the simulation environment using a policy that achieves a success rate of $0.15$. 
The texture of the table surface is randomly sampled (using \cite{cimpoi14describing}) for domain randomization. 
Different lighting conditions are applied to the environment. 
Note that no additional noise is applied to the visual input.
Experiments with 10 objects (randomly selected subset for each seed) and 100 objects are compared.

\textbf{Analysis:}
In \cref{fig:ex4}, we compare the setup without tactile feedback $\mathbf{E}$ 
with the performance of policies that are provided with the sum of force vectors $\mathbf{V}$. Higher-fidelity tactile input did not improve performance in our runs. 
We observe that the baseline performance ($\mathbf{E}$) decreases when the diversity of objects increases, whereas simple tactile input ($\mathbf{V}$) leads to a constant performance even for a larger range of objects.

In \cref{fig:ex4_bar_2}, an MPO actor is trained with the same 100 objects for 900K steps. When testing with the 100 training objects (seen), the simple tactile input (\textbf{V}) brings a significant advantage over the baseline input (\textbf{E}). When testing with 63 unseen objects, both cases with and without tactile sensing experience a performance drop. This is in part because the novel object has fewer grasping poses than the 100 training objects. 
Total force as tactile sensing maintains a relatively high success rate, showing that tactile information aids in generalizing to grasping novel objects.

\subsection{Is the conclusion consistent across sim-2-real?}\label{sec:ex5}
A common way to deploy RL policies on real hardware is through sim-2-real. We set up a hardware experiment according to \cref{sec:ex2} using a Franka Panda robot and a self-developed antipodal gripper equipped with two Minsight sensors as fingertips. We use FoundationPose \cite{wen2024foundationpose} for object pose estimation. The 20\,Hz policy output is interpolated by a joint impedance controller to 1000\,Hz. Modification are necessary for a successful sim-2-real: 
1)~The episode length is increased to 100 steps to overcome the controller tracking discrepancy. 
2)~Action moving average filter and 17\,ms delay are applied during training to mimic the realistic observation and action execution time. 
3)~Friction coefficients and controller parameters are strongly randomized. 
4)~The policy is trained for 1.6M steps. During training, we apply both OU and offset noise to perception. We compare the average success rate for six types of tactile information for three representative objects (\cref{fig:objects} first column)  within 20 tries. 
We also compared three test scenarios as shown in \cref{tab:ex-5}: 
1)~sim: grasping in simulation, using the same setup as \cref{sec:ex2}; 
2)~w/o disturb: grasping in the real world, using a similar setup as \cref{sec:ex2}; 
3)~w/ disturb: similar to 2) but with human disturbance, incl.\ random shifting/reorientation of the object during the test. 
We also report success rates with a programmed gripper closing after the policy rollout for E, B, BK, and MK in the brackets. Real robot test for tactile input \textbf{VK} is excluded due to the limited computational power.

\textbf{Analysis:}
From \cref{tab:ex-5}, we first notice that even with more training steps and longer episode length, the simulation results show a similar tendency as the \cref{fig:ex2} with OU and offset noise. 

Experiments of sum force as tactile input (V) show a stronger sim-2-real performance compared to other setups. 

Experiments for  BK, MK show a weaker sim-2-real performance drop compared to B,M,V. We hypothesize that the inaccurate real robot motion leads to aggressive touching behavior. The denser ``local'' tactile feedback experiences larger distribution shift than the ``global''.

Experiments without tactile sensing (E) showed a near-zero success rate when controlled solely by the policy, in contrast to the higher success rate with programmed closing. This implies that successful grasping is weakly correlated with the gripper motions in the absence of tactile data.

The inversed sim-2-real gap for banana grasping is mainly caused by the increased contact area of the deformable sensor and its high-friction silicone shell.

\section{Discussion}\label{sec:Discussion}

We set out to answer which type of tactile sensing is beneficial for learning to grasp objects with a 2-finger gripper. 
Under realistic effects of imperfect visual perception, 
 we find that tactile information is important for successful grasps.

However, a fine spatial resolution of tactile feedback seems not to be required, which is counterintuitive. 
Reasons might lie in the limited manipulation capabilities of an antipodal 2-finger gripper, as practically no in-hand manipulation is possible.
The orientation of the overall contact forces apparently contains more useful information than spatially distributed taxels with binary and force magnitudes.  

The low success rate in the experiments was primarily due to the use of an antipodal gripper (point-wise contact). Comparative experiments to \cref{sec:ex1} shows that using a parallel gripper (surface contact) with the same training pipeline significantly improved success rates (0.93).

An interesting future direction is to consider blind grasping, which requires active search and accumulation of information. 
We are also wondering how our results would change for multi-finger in-hand manipulation.


\textbf{Limitations:}

Drawing a comprehensive conclusion is challenging due to numerous factors. Comparisons are mostly within a simplified, controlled environment. The sensor model ignores real sensor deformations, and tactile feedback uses rigid-body collision-checking and approximating high-resolution sensing. 
Simulation results may vary based on hyper-parameters and algorithms as seen with SAC and MPO yielding different qualitative outcomes. 
Several factors introduced during the sim-to-real process impact the evaluation of real robot experiments.
Also, we are not using online learning for visual-tactile features (e.g., a modified SLAC \cite{lee2020stochastic}) due to limited computational resources.



\newpage
\bibliographystyle{IEEEtran}
\bibliography{refs}
\end{document}